\documentclass{article}

\usepackage[preprint,nonatbib]{neurips_2024}

\usepackage[utf8]{inputenc} 
\usepackage[T1]{fontenc}    
\usepackage{hyperref}       
\usepackage{url}            
\usepackage{booktabs}       
\usepackage{amsfonts}       
\usepackage{nicefrac}       
\usepackage{microtype}      
\usepackage{xcolor}         

\usepackage{amsmath}
\usepackage{xspace}
\makeatletter
\DeclareRobustCommand\onedot{\futurelet\@let@token\@onedot}
\def\@onedot{\ifx\@let@token.\else.\null\fi\xspace}

\def\eg{\emph{e.g}\onedot} 
\def\ie{\emph{i.e}\onedot}

\def\etal{\emph{et al}\onedot}
\makeatother

\usepackage{diagbox}
\usepackage{svg}
\usepackage{graphicx}
\usepackage{caption}
\usepackage{subcaption}
\usepackage{multirow}
\usepackage{makecell}
\usepackage{utfsym}
\usepackage{color, colortbl}
\usepackage[capitalize]{cleveref}
\crefname{section}{Sec.}{Secs.}
\crefname{section}{Section}{Sections}
\crefname{table}{Table}{Tables}
\crefname{table}{Tab.}{Tabs.}

\definecolor{tabhighlight}{HTML}{e5e5e5}

\title{Towards Efficient Vision-Language Tuning: More Information Density, More Generalizability}

\makeatletter
\newcommand{\myfnsymbol}[1]{%
  \expandafter\@myfnsymbol\csname c@#1\endcsname
}
\newcommand{\@myfnsymbol}[1]{%
  \ifcase #1
  \or 1
  \or 2
  \or \TextOrMath{\textasteriskcentered}{*}
  \or \TextOrMath{\textdagger}{\dagger}
  \fi
}
\newcommand{\affiliationA}{\@myfnsymbol{1}}
\newcommand{\affiliationB}{\@myfnsymbol{2}}
\newcommand{\equalcontributor}{\@myfnsymbol{3}}
\newcommand{\correspondingA}{\@myfnsymbol{4}}
\makeatother

\author{Tianxiang Hao$^{1,2}$\textsuperscript{\equalcontributor}, Mengyao Lyu$^{1,2}$\textsuperscript{\equalcontributor}, Hui Chen$^{2}$\textsuperscript{\correspondingA}, Sicheng Zhao$^{2}$, \\
\textbf{Xiaohan Ding$^{3}$}\textbf{, Jungong Han}$^{4}$\textbf{, Guiguang Ding}$^{1,2}$\textsuperscript{\correspondingA}\\
$^{1}$School of Software, Tsinghua University \, $^{2}$BNRist\\
$^{3}$Bytedance \, $^{4}$Computer Science Department, University of Sheffied\\
\texttt{\{beyondhtx,jichenhui2012,jungonghan77\}@gmail.com} \\
\texttt{mengyao.lyu@outlook.com, \{schzhao,dinggg\}@tsinghua.edu.cn}
}

\begin{document}

\renewcommand{\thefootnote}{\myfnsymbol{footnote}}
\maketitle
\footnotetext[3]{Equal contribution}%
\footnotetext[4]{Corresponding authors}%

\setcounter{footnote}{0}
\renewcommand{\thefootnote}{\fnsymbol{footnote}}

\maketitle
\begin{abstract}
  With the advancement of large pre-trained vision-language models, effectively transferring the knowledge embedded within these foundational models to downstream tasks has become a pivotal topic, particularly in data-scarce environments. Recently, parameter-efficient fine-tuning approaches, especially prompt tuning, have garnered considerable attention. To better understand the nature of prompt tuning, we propose the concept of ``Information Density'' (ID) to indicate whether a matrix strongly belongs to certain feature spaces rather than being evenly distributed across various feature spaces. We suppose a higher ID with strong bias across some feature spaces naturally leads to excellent robustness and stability. Our research, inspired by the observation that generalizability is closely linked to the information density of the prompt matrix, introduces the Dense Information Prompt (DIP). DIP aims to enhance information density to improve generalization. Furthermore, DIP significantly reduces the number of tunable parameters and the requisite storage space, making it particularly advantageous in resource-constrained settings. Comprehensive experiments substantiate the superiority of DIP. Notably, DIP surpasses the latest state-of-the-art methods by a substantial margin with an exceptionally small parameter count. Across a range of tasks spanning 11 datasets, DIP improves the average downstream accuracy of classic prompt tuning by up to 5.76\% using merely 0.5K parameters.
\end{abstract}

\section{Introduction}
\label{sec:intro}

In recent years, vision-languages models~\cite{radford2021learning,jia2021align} have achieved tremendous success. Representative models like CLIP~\cite{radford2021learning} are first pre-trained on a huge number of text-image pairs on the web to align textual and visual features, and then can be tuned and used for various downstream tasks.

However, traditional fine-tuning is not a good choice to adapt vision-language models. Simply fine-tuning all the parameters can easily cause the model to overfit because the huge number of parameters bring redundant non-essential information. The huge training and storage cost is also an intractable problem. In the context of our study, we introduce the concept of ``information density''. Much like the rank of a matrix in linear algebra, which represents the maximum number of linearly independent rows or columns in the matrix, ``information density'' represents the maximum amount of essential and non-redundant information that the model can extract from the downstream task. Just as a matrix with a higher rank possesses more unique information, a model with high information density can acquire more essential and general information from the downstream task by fewer parameters, even with a smaller dataset. Our goal is increasing the information density and thus using the fewest but most essential parameters to finish generalization, without causing catastrophic forgetting or overfitting to the small dataset.

As the concept of information density can be functionally analogous to the rank of a matrix, we decided to use properties related to rank to quantify information density. Specifically, we take a full-rank matrix and decompose it using Singular Value Decomposition (SVD) to obtain a matrix of singular values, using the properties of these singular values to define and quantize information density. We found that this definition of information density is highly correlated with the model's generalization performance (Spearman correlation coefficient $>$ 0.9). Therefore, we propose DIP, aiming to enhance the model's generalization ability by increasing information density. Additionally, due to the increased information carried by each parameter unit, our approach can significantly reduce the required number of parameters.

In \cref{sec:motivation}, we will show our finding about the strong correlation between generalization capability and information density of the prompt matrix in \cref{fig:motivation}. Inspired by such observation, we propose Dense Information Prompt (DIP) for effective and efficient adaptation, well adapting models under an extremely small number of parameters where nobody has explored before. There are several good advantages of DIP:

\begin{itemize}
    \item \textbf{Efficiency and Effectiveness} To our knowledge, we are the first one to reach comparable or even better performance with state-of-the-art methods using such an extremely small number of parameters, \ie 516. 
    \item \textbf{Simplicity} Replacing the classic prompt by DIP just needs to modify several lines of code. No special loss nor extra module is introduced.
    \item \textbf{Robustness} DIP is relatively capable of anti-disturbance. As in \cref{tab:domain_generalization}, DIP could maximally reserve its knowledge from domain shift. 
    \item \textbf{Plug and Play} For any classic model, DIP only replaces the prompts, enabling us to plug DIP into most of the existing methods fruitfully.
\end{itemize}

In summary, we conclude our contributions as follows: 

\vspace{-1mm}
\begin{itemize}
\setlength{\topsep}{0pt}
\setlength{\parsep}{0pt}
\setlength{\itemsep}{0pt}
\setlength{\partopsep}{0pt}
\setlength{\parskip}{0pt}
    \item We propose a new concept ``information density'', give its definition and well-quantize the concept. We further find the strong correlation between generalizability and information density in \cref{fig:motivation}, and thus propose to use DIP to increase information density for better generalization capability.
    \item We propose a novel initialization method and integrate a lightweight regularization module to further improve the performance of Dense Information Prompt tuning without introducing any extra parameters and inference cost.
    \item We are the first to explore how to effectively adapt vision-language models using an extremely small number of parameters, \ie ~0.5K.
    \item We conduct extensive experiments and show the fantastic effectiveness and efficiency of DIP. In base-to-new generalization, domain generalization, cross-dataset transfer and few-shot learning settings, DIP consistently reaches very competitive results, surpassing many state-of-the-art tuning methods though we use fewer parameters.
\end{itemize}

\section{Related Works}
\label{sec:related}

\subsection{Vision-Language Models} 
Recently, large-scale vision-language models have shown very competitive performance in various tasks. Classic works~\cite{radford2021learning,jia2021align,zhai2022lit,yao2022filip,yuan2021florence} learn the multi-modal representation by a self-supervised manner on a large amount of image-text pairs. The representative work CLIP~\cite{radford2021learning} is a milestone, which aligns the vision representation and language representation by contrastive learning and shows excellent performance.
A well-trained vision-language model is a great treasure, which could largely facilitate the development of many fields. There have been successful applications of such strong models on few-shot recognition~\cite{zhou2022conditional,zhou2021coop},
detection~\cite{rasheed2022bridging,maaz2022class,feng2022promptdet,zang2022open} and segmentation~\cite{li2022languagedriven,rao2022denseclip,ding2022decoupling,luddecke2022image}. For video data, there are also works on video classification~\cite{qian2022multimodal} and video understanding~\cite{ju2022promptingvideounderstand}.

\subsection{Prompt Tuning} Prompt tuning is one of the most popular methods to tune models in downstream tasks with excellent efficiency. Originating from natural language processing, prompts are first introduced as a fixed template~\cite{schick2020pet}, \eg \textit{a photo of a \_}, which is hand-crafted and fixed. Later, a series of methods~\cite{li2021prefix,lester2021power,liu2021p,shin2020autoprompt,liu2021pre,jiang2021can} are proposed to make such prompts tunable and be optimized during adaptation. Prompt tuning could adaptively narrow the gap between pre-trained representations and downstream tasks, significantly facilitating the fine-tuning process. Representative prompt tuning methods would add tunable virtual tokens, \ie prompts, along with the semantic tokens as inputs of the model. All of the tokens are processed together to get text embeddings first and then sent to the feature encoder.  Witnessing the success of prompting language models, researchers design prompts~\cite{jia2022vpt,zhang2022noah} for visual models in a similar way. In vision-language field, there are several explorations as well. Bahng \etal~\cite{bahng2022visual} adopts prompt tuning merely on the image encoder. CoOp~\cite{zhou2021coop} uses tunable text prompts to replace the fixed template in CLIP~\cite{radford2021learning}. CoCoOp~\cite{zhou2022conditional} utilizes image feature to instruct the optimization of the tunable text prompts in CoOp. ~\cite{khattak2023maple,lee2023rpo} simultaneously optimize image and text prompts and establish extra connections between different modals. ~\cite{khattak2023promptsrc,yao2023kgcoop,bulat2023lasp,zheng2023regularized,hao2024quantized} integrate strong regularization modules or losses into prompt tuning to diminish the overfitting and catastrophic forgetting problem. For better downstream accuracy, researchers design more and more complicated methods, accompanied by inefficiency. To solve the problem, we propose Dense Information Prompt (DIP) to take the place of classic prompts, which can largely decrease the number of tunable parameters and further enhance the model's generalization ability. Notice that though becoming more complex, existing methods are still refined on a common fundamental basis, \ie prompt tuning. Such a common basis guarantees that DIP could be easily and smoothly integrated into most of the off-the-shelf methods besides individually applied. Besides prompt-based methods, there are also many other works to acquire storage efficiency~\cite{hao2023consolidator,houlsby2019parameter,hu2021lora,lian2022ssf,zhang2022noah,chen2022adaptformer}, inference efficiency~\cite{chen2022mtp,wang2023cait,bolya2023tome,hao2023manipulating,ding2021resrep,ding2019centripetal,chen2023diffrate,shen2024tempme,xiong2024pyra} and data efficiency~\cite{wang2023mhpl,lyu2024lftl} during downstream tuning.

\section{Relationship between Information Density and Generalizability}
\label{sec:motivation}

In this section, we will start from reviewing a classic prompt tuning pipeline CoOp~\cite{zhou2021coop} on CLIP in \cref{subsec:clip}, and propose a new concept ``Information Density'' and analyze its relationship with generalizability in \cref{subsec:information_density}.

\subsection{A Review of Prompt Tuning for CLIP}
\label{subsec:clip}
CLIP consists of a text encoder $\mathcal{L}$ and an image encoder $\mathcal{V}$. Typically, $\mathcal{L}$ is a language transformer, while $\mathcal{V}$ can be a convolutional neural network or a vision transformer. In this paper, we follows ~\cite{zhou2021coop,zhou2022conditional} to use a ViT-B/16~\cite{dosovitskiy2020image} as the image encoder $\mathcal{V}$ unless specifically mentioned. We start by making a review of how to prompt a CLIP for prediction in the following paragraphs.
\\
\textbf{Text Encoder} Suppose there are $M$ layers in the text encoder. For $k$-th layer $\mathcal{L}_{k}$, the inputs are a series of prompt tokens $P^{l}_{k-1}$ and a \textit{[CLS]} token $c^{l}_{k-1}$, and the outputs are $P^{l}_{k}$ and $c^{l}_{k}$. The inputs of the first layer $P^{l}_{0}$ and $c^{l}_{0}$ are exactly the word embeddings of the prompts along with the label, \eg ``\textit{A photo of a [CLS]}'' or just some randomly initialized vectors. Formally, we have $P^{l}_{k}\in \mathbb{R}^{n^{l}\times d^{l}}$ and $c^{l}_{k}\in \mathbb{R}^{d^{l}}$, where $n^{l}$ denotes the text prompts' length and $d^{l}$ denotes the dimension of word embedding. $\forall 1\leq k \leq M$, we have $[P^{l}_{k},c^{l}_{k}]=\mathcal{L}_{k}([P^{l}_{k-1},c^{l}_{k-1}])$.


The output feature of the text encoder $f^l\in \mathbb{R}^{d^v}$, where $d^v$ is the dimension of the visual feature space, is generated by projecting the \textit{[CLS]} token of the last layer to the visual space by a linear transformation, \ie $f^{l}={\rm Proj}(c^{l}_{M})$.
\\
\textbf{Image Encoder} Suppose there are $N$ layers in the image encoder. For $k$-th layer $\mathcal{V}_{k}$, the inputs are image patch tokens $I_{k-1}$, a classification token $c^{v}_{k-1}$ and prompt tokens $P^{v}_{k-1}$, and the outputs are $I_{k}$, $c^{v}_{k}$ and $P^{v}_{k}$. The inputs of the first layer $I_{0}$ and $c^{v}_{0}$ are exactly the patch embeddings of the image and pre-trained class token. $P^{v}_{0}$ is randomly initialized in general. Formally, we have $I_{k}\in \mathbb{R}^{p\times d^v}$, $c^{v}_{k}\in \mathbb{R}^{d^v}$ and $P^{v}_{k}\in \mathbb{R}^{n^{v}\times d^{v}}$, where $p$ denotes the number of image patches and $d^{v}$ denotes the dimension of visual embedding. $\forall 1\leq k \leq N$, $[P^{v}_{k},c^{v}_{k},I_{k}]=\mathcal{V}_{k}([P^{v}_{k-1},c^{v}_{k-1},I_{k-1}])$. The output feature of the image encoder is $f^{v}=c^{v}_{N}$.


\textbf{Prediction} CLIP can be used for image classification. Suppose there are $C$ classes, and $\{f^l_{c}\}_{c=1}^{C}$ are the corresponding text features. Label $y$'s probability is $p(y|f^v)=\frac{\mbox{exp}(\mbox{sim}(f^v,f^{l}_{y})/\tau)}{\sum_{c=1}^{C}\mbox{exp}(\mbox{sim}(f^v,f^{l}_{c})/\tau)}$ where sim$(\cdot,\cdot)$ denotes cosine similarity function and $\tau$ is temperature. The final prediction is $\hat{z}=\mathop{\arg\max}\limits_{1\leq y\leq C}(p(y|f^v))$.


It is worth noting that some researchers adopt a deeper manner~\cite{jia2022vpt,khattak2023maple} to organize the prompts. They directly add and tune the prompt in each layer in the feature encoder, instead of inheriting the output prompt calculated by the last encoder, \ie a forward pass becomes $[\_,c^{l}_{k}]=\mathcal{L}_{k}([P^{l}_{k-1},c^{l}_{k-1}])$ and $[\_,c^{v}_{k},I_{k}]=\mathcal{V}_{k}([P^{v}_{k-1},c^{v}_{k-1},I_{k-1}])$. Each $P^l$/$P^v$ contains tunable parameters.

\begin{figure}[tb]
  \centering
    \setlength{\belowcaptionskip}{-0.2cm}
    \includegraphics[width=0.8\linewidth]{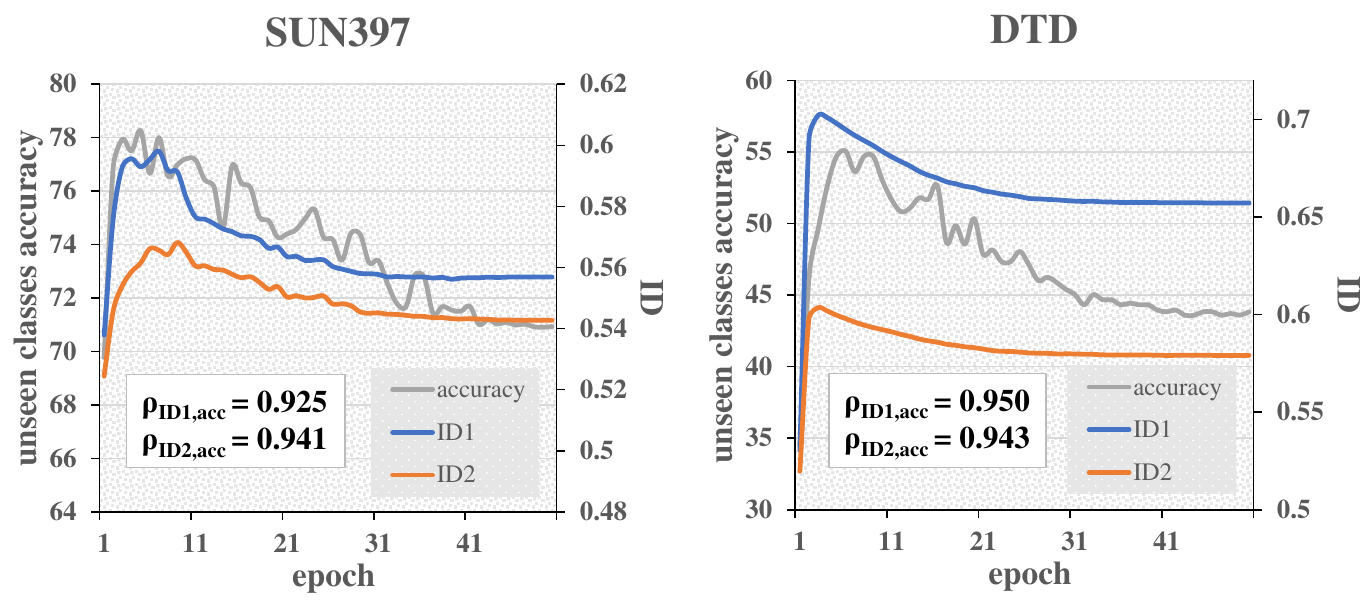}
  \caption{Relationship between generalizability represented by the test accuracy on unseen classes during training and Information Density (ID). When generalizability increases, ID also increases. The Spearman correlation coefficient $\rho$ between generalizability and ID1/ID2 is very high, \ie $\geq$ 0.9. }
  \label{fig:motivation}
\end{figure}

\subsection{Information Density in Prompt Tuning}
\label{subsec:information_density}
Here, we first provide precise definitions for ``information density'' to clearly convey our motivations. For typical parameter matrix like prompts $P\in \mathbb{R}^{n\times d}$(assume $n<d$), we can always rewrite such matrix into a combination of several orthogonal bases with different weights by singular value decomposition (SVD). Formally, $P=U\Sigma V^T=\sum_{i=1}^{d}\sigma_{i}u_{i}v_i^T$. Each $u_iv_i^T$ can span a unique feature space, and $P$ is a linear combination of these features. Typically, $\{\sigma_i\}_{i=1}^n$ are arranged in descending order.

To help readers better understand the concept of information density, let's draw an analogy using images from nature. A real image always has one or a few very prominent features and almost never contains an even mix of various odd features. In an extreme case, if an image truly exhibits isotropic characteristics, it would simply mean that its content is almost entirely noise and lacks clear meaning. Reflecting this back to the matrix decomposition we discussed earlier, a good image tends to have several significantly larger singular values. The features of the image can largely be expressed by the feature space behind these prominent singular values. 
Therefore, from the matrix decomposition expression, the differences among the singular values $\{\sigma_i\}_{i=1}^{n}$ are significant, indicating that the information is concentrated in a few feature spaces. In other words, the information density is higher. Thus, to quantize such property, we define $k$-th order ``Information Density (ID)'' as follows: ID$k=\frac{\sum_{i=1}^k\sigma_i}{\sum_{i=1}^n\sigma_i}$. In other words, ID$k$ is the proportion of the largest $k$ singular values among all the singular values. Greater information density represents more robust and stable intrinsic features, meaning they are less likely to be affected by external disturbances and have stronger anti-interference capabilities.

Returning to our initial discussion, a core contribution of this paper is the hypothesis and verification that the parameter matrices, like prompts, follow the same ID-related laws during fine-tuning in downstream CLIP models. As is well-known, in transfer learning, the transfer of knowledge in a model depends on the updating of parameters. For CLIP models, the optimal solution is prompt tuning~\cite{zhou2021coop,zhou2022conditional,khattak2023maple,zhu2023prograd}. We first hypothesize that the information density of the prompt matrix also represents its robustness and the strength of its intrinsic features. Therefore, greater information density should theoretically result in better generalizability throughout the prompt tuning process.

To verify this hypothesis, we conducted an experiment using a classic method, CoOp~\cite{zhou2021coop}. During training, we masked half of the classes, using data from only the other half for training, and performed singular value decomposition on the prompt matrices during the process. As shown in \cref{fig:motivation}, for better visualization, ID1 is scaled up to 2x to be put under the same right axis with ID2. Unseen classes accuracy is improved in the first few iterations, but it starts dropping later, indicating overfitting and catastrophic forgetting. Importantly, the fluctuation trend of unseen classes accuracy is highly consistent with ID. We compute the Spearman correlation coefficient between unseen classes accuracy and ID1/ID2 in \cref{fig:motivation}. Clearly, the first-order and second-order information densities of CoOp on the SUN397~\cite{xiao2010sun} and DTD~\cite{cimpoi2014describing} datasets exhibit a very strong correlation with the accuracy on unseen classes (\ie, generalizability), with Spearman correlation coefficients greater than 0.9. This demonstrates that our hypothesis is correct.

In the following \cref{sec:method}, we will show how we can leverage such correlation between generalizability and ID to boost CLIP's downstream performance.

\section{Methodology}
\label{sec:method}
\subsection{Dense Information Prompt}
\label{subsec:rep-prompts}
\subsubsection{Algorithms for increasing information density}

In this part, we aim to find an effective algorithm to increase information density to enhance CLIP's downstream generalization performance. Motivated by the observation in \cref{sec:motivation}, we propose three optional algorithms. 

\textbf{Alg. 1: Do SVD and directly optimize information density as an training objective. } Specifically, we add corresponding expression to the loss item after decomposing prompt matrix by SVD. Suppose the original cross entropy loss is $\mathcal{L}_{CE}$ and we want to maximize ID$k$ ($1\leq k\leq n)$, the current loss is $\mathcal{L}=\mathcal{L}_{CE}-\lambda \mbox{ID}k$, where $\lambda$ is a positive hyper-parameter.

\textbf{Alg. 2: Do SVD and apply order-related penalty on the singular values.} Specifically, we add regularization term to our optimization objective. Suppose the original cross entropy loss is $\mathcal{L}_{CE}$ and we want to maximize ID$k$ ($1\leq k\leq n)$, the current loss is $\mathcal{L}=\mathcal{L}_{CE}+\lambda \sum_{i=n-k+1}^{n}i||\sigma_i||$.

\textbf{Alg. 3: Approximate the original matrix by the product of two small matrices.} Formally, given a typical prompt matrix $P\in \mathbb{R}^{n\times d}$, we take inspiration from LoRA~\cite{hu2021lora} to create two low-dimensional matrices and use their product as an approximated equivalent prompt. Suppose we want to maximize ID$r$ ($1\leq r\leq n)$, we first randomly initialize $P_{A}\in \mathbb{R}^{n\times r}$ and $P_{B}\in \mathbb{R}^{r\times d}$. The low-rank approximated prompt $P_{lr}=P_{A}P_{B}$ is with the same shape with $P$, and thus can participate in the training.

We do a quick verification on several datasets used in the base-to-new generalization setting and show the results in \cref{tab:verify}. Clearly, from Alg. 1 to Alg. 3, the intensity of constraint is becoming stronger, resulting in more accuracy loss on base classes while enjoying more accuracy gain on new classes. In addition to the good performance, Alg. 3 has another advantages: it's quite simple and easy to implement with faster training speed. Therefore, we choose Alg. 3 to implement our DIP. However, unlike LoRA who merely uses a similar low-rank product as a supplementary addition for the output of main branch, Alg. 3 directly outputs the product for usage, leading to some unique difficulties and challenges. We will show the problems behind such approximation and our solutions in the next sections.

\begin{table}[t]
    \centering
    \scriptsize
    \setlength{\abovecaptionskip}{-0.01cm}
    \setlength{\belowcaptionskip}{-0.05cm}
    \caption{A quick check for different implementations of DIP.}
    \setlength{\tabcolsep}{2.7mm}{
    \renewcommand{\arraystretch}{1.2}
    \begin{tabular}{|c|c|c|c|}
        \hline
         & base & new & H \\
         \hline
         w/o any & 82.69 & 63.22 & 71.66\\
        \hline
        Alg. 1 & 81.21 & 67.58 & 73.77 \\
        \hline
        Alg. 2 & 80.39 & 69.85 & 74.75\\
        \hline
        Alg. 3 & 79.91 & 71.48 & \textbf{75.46}\\
        \hline
    \end{tabular}
    }
    \label{tab:verify}
\end{table}


\subsubsection{Special Initialization}

In the field of tuning vision-language models, existing works have confirmed that the initialization method of prompts is quite important. For example, ~\cite{zhou2022conditional,zhou2021coop} adopt a hand-crafted template as the initial point of the text prompts, and ~\cite{lee2023rpo} copies the parameters in the text or image class token to initialize the prompts of the corresponding branch. If we just do a random initialization, the overall performance of such existing methods would drop severely. See \cref{subsec:analysis} for more details. In other words, it would be helpful if we could take advantage of a good initial point.

The problem lies in the fact that artificially designed initialization $P_{init}$ is almost certain to be full rank, \ie $\mbox{rank}(P_{init})=min(n,d)$, and our current low-rank prompt $P_{lr}=P_{A}P_{B}$ could only express the matrices whose rank is less or equal than $r$. Since $r<min(n,d)$, it is impossible to directly initialize $P_{A}$ and $P_{B}$ by a given $P_{init}$. 

\begin{figure}[tb]
  \centering
  \begin{subfigure}{0.6\linewidth}
    \includegraphics[width=0.95\linewidth]{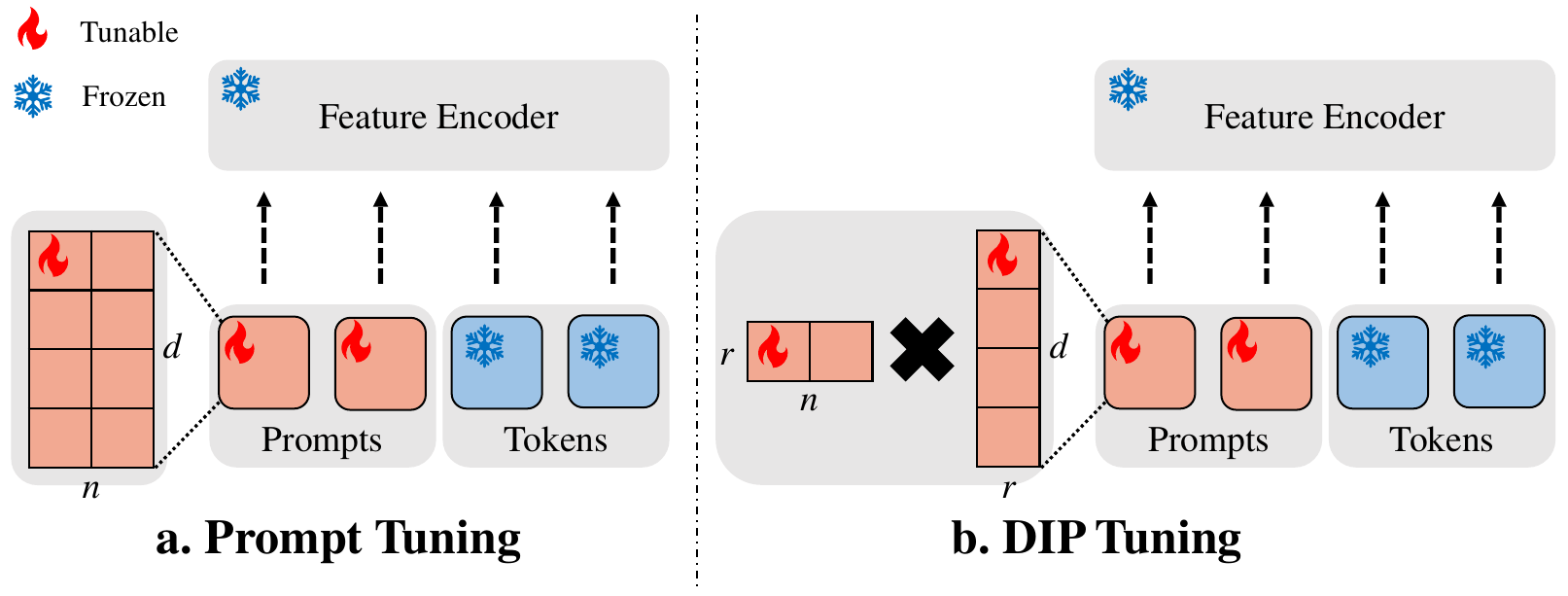}
\caption{Comparison between DIP Tuning and Prompt Tuning.}
    \label{fig:compare_pt_DIP}
  \end{subfigure}
  \begin{subfigure}{0.38\linewidth}
    \includegraphics[width=0.95\textwidth]{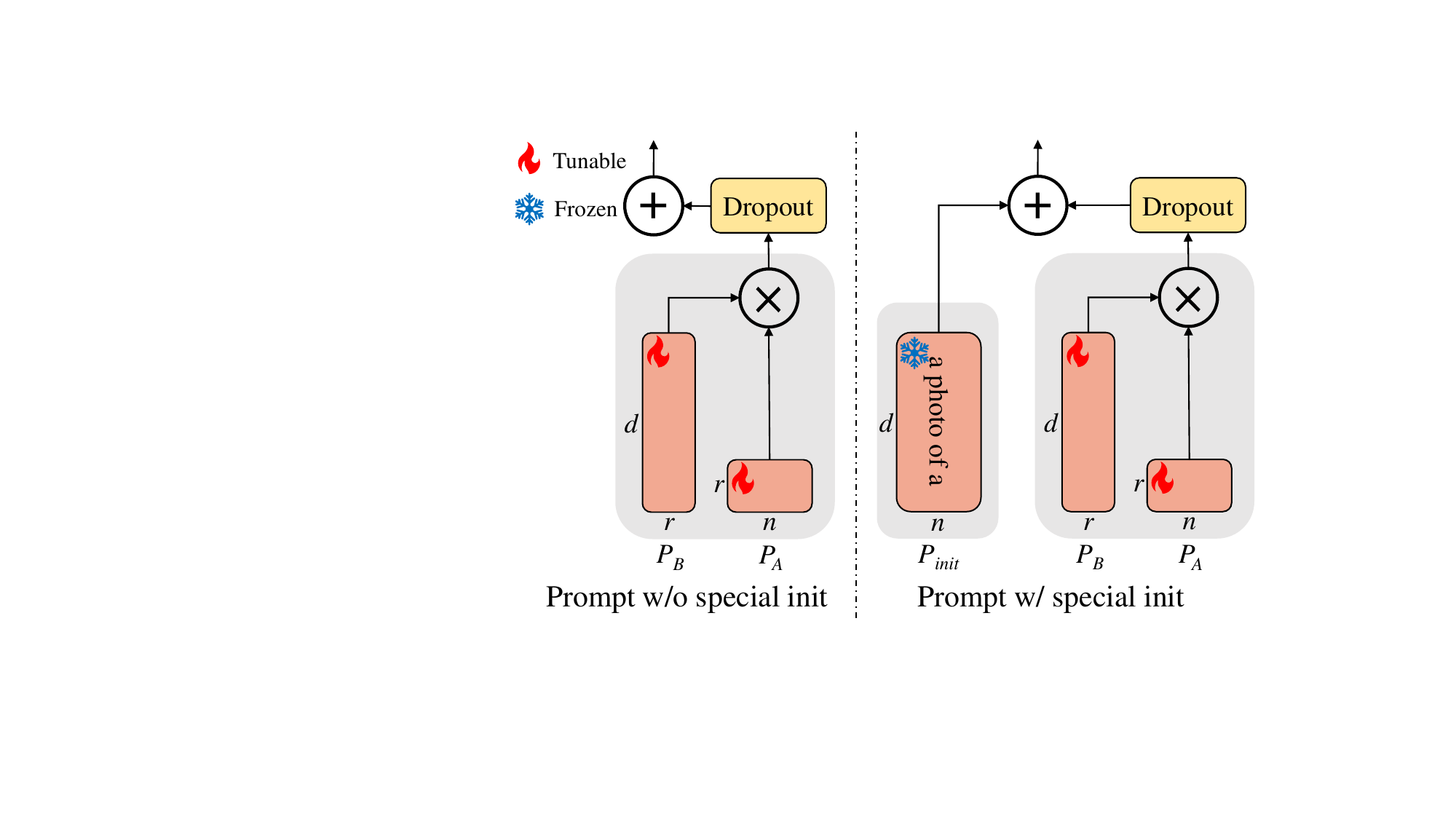}
\caption{DIP enjoys special initialization.}
\label{fig:fullrank_init}
  \end{subfigure}
  \setlength{\abovecaptionskip}{-0.01cm}
  \setlength{\belowcaptionskip}{-0.5cm}
  \caption{To switch from classic prompt tuning and to DIP tuning, just replace the ordinary prompts with DIPs. DIP introduces two small parameter matrices in shape $[n,r]$ and $[r,d]$ separately, and uses their product as an equivalent prompt in shape $[n,d]$. For the prompts with special initialization, \eg a hand-crafted template \textit{``a photo of a''} for the text prompts, we introduce a concurrent full-rank prompt branch along with the proposed low-rank prompts. By turning off the gradient of the newly added branch, we start training from a promising initial point, and the total number of tunable parameters or stored parameters will not increase as well. The Dropout layer could effectively regularize the update of low-rank prompts and alleviate overfitting and catastrophic forgetting. Dropout is a lightweight non-parametric layer and turns out to be an Identity layer in inference, resulting in negligible cost.}
  \label{fig:DIP_overfiew}
\end{figure}

To solve such a problem, we add a concurrent branch of full-rank prompts $P_{fr}\in \mathbb{R}^{n\times d}$ along with the proposed low-rank prompts as shown in \cref{fig:fullrank_init}. Naturally, $P_{fr}=P_{init}$. And we randomly sample $P_{A}$/$P_{B}$ from a Gaussian distribution in which $\mu=0$ and $\sigma \rightarrow 0$. A small $\sigma$ here could avoid constant initialization and enrich the update paths. Notably, $P_{fr}$ is kept frozen during the whole adaptation, and thus it would not increase the number of tunable or stored parameters.

\subsubsection{Regularization}

As discussed in \cref{sec:related}, existing works have shown that proper regularization would significantly improve the generalization ability. Therefore, to alleviate overfitting and catastrophic forgetting, we put a Dropout layer with drop ratio $p$ after the low-rank branch as displayed in \cref{fig:fullrank_init}.

Therefore, the input prompt of the feature encoder is $P=P_{fr}+\mbox{Dropout}(P_{lr},p)$.
Finally, we have $P=P_{init}+\mbox{Dropout}(P_{A}P_{B},p)$.


\subsection{Efficiency Analysis}

The whole adaptation process of a pre-trained vision-language model can be divided into three parts: training, storage and inference. In this subsection, we will analyze the efficiency of DIP in each part separately.
\\
\textbf{Training} In the training phase, the number of tunable parameters in total is $r(n+d)$. Compared with classic prompt tuning which has $nd$ tunable parameters, we can largely reduce the trainable parameters by choosing a small $r$ satisfying $r<<min(n,d)$. Such reduction could help us train the model faster and use less memory during training than those delicate-designed methods with massive complicated structures and update rules. 
\\
\textbf{Storage} After training, we just store $P_{A}$ and $P_{B}$ onto the disk for every downstream task. Similar to the training period, the number of stored parameters is $r(n+d)$. In contrast, classic prompt tuning stores a much larger set of $nd$ parameters. As a result, we can save a fairly large amount of storage space when there are lots of tasks to be adapted.
\\
\textbf{Inference} Before inference, we first load $P_{init}$, $P_{A}$ and $P_{B}$ from disk to memory. Noticing that Dropout is exactly an identity layer in the inference mode, we could pre-calculate the equivalent $P$ by $P=P_{init}+P_{A}P_{B}$ and just keep $P$ in the memory. For inference, we directly use $P$ as the input prompts, and thus the inference cost is the same as classic prompt tuning. Some existing methods add complex bridges between the isolated parameters to earn extra improvements, \eg CoCoOp~\cite{zhou2022conditional}. There ain’t no such thing as a free lunch. They would face slower speed and huge memory occupation in inference time.

\section{Experiments}
\label{sec:experiments}

To verify the effectiveness of the proposed method, we evaluate our method and make comparisons with the latest state-of-the-art methods in terms of the following settings in a wide range: base-to-new generalization, domain generalization, cross-dataset transfer and few-shot learning.

For more experimental details, please refer to the Appendix.

\begin{table}[tb]
    \scriptsize
    \setlength{\belowcaptionskip}{-0.05cm}
    \setlength{\tabcolsep}{1.5mm}
    \caption{Comparisons with latest methods in base-to-new generalization. H: harmonic mean~\cite{xian2017zero}. In the lightweight methods, DIP+CoOp outperforms all the other competitors on new and harmonic mean accuracy. While in the heavy methods, DIP+MaPLe surpasses MaPLe~\cite{khattak2023maple} by a clear margin, and is strongest on base, new and harmonic mean accuracy, even with around only $\frac{1}{5}$ parameters compared with MaPLe.}
    \label{tab:results_base_to_new_generalization}
    
    \centering
    \begin{tabular}{lccc|c}
    \toprule
    Method & \#params & Base & New & H \\
    \midrule
    \textbf{Lightweight}  & & & & \\
    CLIP~\cite{radford2021learning} & 0K & 69.34 & 74.22 & 71.70 \\
    CoOp~\cite{zhou2021coop} & 2.1K & 82.69 & 63.22 & 71.66 \\
    CoCoOp~\cite{zhou2022conditional} & 35.4K & 80.47 & 71.69 & 75.83 \\
    Adapter~\cite{gao2021clip-adapter} & 525.5K & 82.62 & 70.97 & 76.35 \\
    LoRA~\cite{hu2021lora} & 129.0K & \textbf{84.30} & 67.33 & 74.86 \\
    ProGrad~\cite{zhu2023prograd} & 8.2K & 82.79 & 68.55 & 75.00 \\
    \rowcolor{tabhighlight}
    DIP+CoOp & 0.5K & 80.32 & \textbf{74.73} &\textbf{77.42} \\
    \hline
    \textbf{Heavy}& & & & \\
    MaPLe~\cite{khattak2023maple} & 3548K & 82.28 & 75.14 & 78.55 \\
    \rowcolor{tabhighlight}
    DIP+MaPLe & 741.9K & \textbf{83.17} & \textbf{75.43} &\textbf{79.11} \\
    \bottomrule
    \end{tabular}
\end{table}

\subsection{Main Results}
\label{subsec:main_result}
\subsubsection{Base-to-new Generalization}

The average results over 11 datasets are shown in \cref{tab:results_base_to_new_generalization}. For complete results, please refer to the Appendix. Overall, DIP shows the strongest performance, and the superiority mainly relies on the improvement of new classes. In other words, DIP largely improves the generalization ability of CLIP. In particular, compared with our direct baseline CoOp, DIP gets $10.95$\% accuracy gain on the new classes and $2.83$\% accuracy drop on the base classes. By adding DIP to the prompts as well as the linear layers of MaPLe, we earn a excellent \textbf{0.56\%} performance gain while reducing its parameters to $0.2\times$ of the original parameters. Notably, compared with the latest lightweight method ProGrad, even if DIP only uses $<$20x of its parameters,  DIP still outperforms it by a clear margin, which clearly demonstrates the efficiency and effectiveness.

\begin{table}[tb]
\scriptsize
\setlength{\tabcolsep}{0.8mm}
\centering
\setlength{\belowcaptionskip}{-0.05cm}
\caption{Comparisons with latest methods in domain generalization after tuned on ImageNet. DIP+CoOp shows excellent robustness when dealing with domain shift. }
\label{tab:domain_generalization}
\begin{tabular}{lc|p{1.6cm}<{\centering}p{1.6cm}<{\centering}p{1.8cm}<{\centering}p{1.6cm}<{\centering}|c}
\toprule
& \#params &  IN-V2 & IN-Sketch & IN-Adversarial & IN-Rendition & Average \\
\midrule
CLIP  & 0K & 60.83 & 46.15 & 47.77 & 73.96 & 57.18 \\
CoOp  & 2.1K & 64.20 & 47.99 & 49.71 & 75.21 & 59.28 \\
CoCoOp  & 35.4K & 64.07 & 48.75 & 50.63 & 76.18 & 59.91 \\
Adapter & 525.5K & 62.53 & 47.67 & 49.17 & 75.42 & 58.70 \\
LoRA & 129.0K & 62.37 & 42.43 & 38.40 & 68.97 & 53.04 \\
ProGrad  & 8.2K & \textbf{64.73} & 47.61 & 49.39 & 74.58 & 59.07 \\
\rowcolor{tabhighlight}
DIP+CoOp & 0.5K & 63.95 & \textbf{49.07} & \textbf{50.97} & \textbf{77.19} & \textbf{60.30} \\
\bottomrule
\end{tabular}
\end{table}

\subsubsection{Domain Generalization}
Then we follow CoCoOp~\cite{zhou2022conditional} to use ImageNet, ImageNet-A, ImageNet-R, ImageNet-v2, and ImageNet-S to run domain generalization experiments to verify the robustness of DIP (DIP = DIP + CoOp, and the same applies thereafter, unless otherwise specified.). Shown in \cref{tab:domain_generalization}, on target datasets, DIP leads to better average accuracy compared with the latest methods, largely outperforming CLIP, CoOp, CLIP-Adapter and ProGrad with much fewer parameters.

\begin{table}[tb]
\scriptsize
\centering
\setlength{\belowcaptionskip}{-0.05cm}
\caption{Results in the cross-dataset transfer setting. DIP+CoOp gives the highest accuracy on 6 of 10 datasets, and slightly outperforms CoCoOp on average. Such result well demonstrates that DIP could maximally extract general and data-agnostic knowledge from given images.}
\label{tab:crossdataset}
\begin{tabular}{lc|cccccccccc|c}
\toprule
& \rotatebox{90}{\# params} & \rotatebox{90}{Caltech101} & \rotatebox{90}{Pets} & \rotatebox{90}{Cars} & \rotatebox{90}{Flowers} & \rotatebox{90}{Food101} & \rotatebox{90}{Aircraft} & \rotatebox{90}{Sun397} & \rotatebox{90}{DTD} & \rotatebox{90}{EuroSAT} & \rotatebox{90}{UCF101} & \rotatebox{90}{Average}\\
\midrule
CoOp & 2.1K & 93.70 & 89.14 & 64.51 & 68.71 & 85.30 & 18.47 & 64.15 & 41.92 & 46.39 & 66.55 & 63.88\\
CoCoOp & 35.4K & \textbf{94.43} & 90.14 & 65.32 & \textbf{71.88} & 86.06 & 22.94 & 67.36 & 45.73 & 45.37 & 68.21 & 65.74\\
Adapter & 525.5K & 93.43 & 88.87 & 64.40 & 70.27 & 85.63 & \textbf{24.67} & 65.80 & 44.90 & \textbf{47.70} & 66.00 & 65.17\\
\rowcolor{tabhighlight}
DIP+CoOp & 0.5K & 94.20 &  \textbf{90.50} & \textbf{67.17} & 71.27 & \textbf{86.07} & 23.83 & \textbf{67.60} & \textbf{46.73} & 42.10 &\textbf{68.93}  & \textbf{65.84}\\
\bottomrule
\end{tabular}
\end{table}

\subsubsection{Cross-dataset Transfer}
Finally, we follow CoCoOp~\cite{zhou2022conditional} to conduct cross-dataset transfer evaluation. Results are shown in \cref{tab:crossdataset}. Concentrating too much on the current dataset will absolutely cause overfitting and catastrophic forgetting problems, and finally lead to a severe drop in the performance on those unseen datasets. In this setting, DIP wins on 6 of 10 datasets and its average accuracy is also slightly better than the best competitor CoCoOp. Such result well demonstrates that DIP could maximally extract general and data-agnostic knowledge from given images compared with other prompt-based methods. Considering the huge difference in the parameter numbers, we could summarize that DIP is still the better choice.

\subsubsection{Few-shot Learning}
\begin{figure}[tb]
\centering
\begin{minipage}[t]{0.42\textwidth}
\centering
    \includegraphics[width=0.95\linewidth]{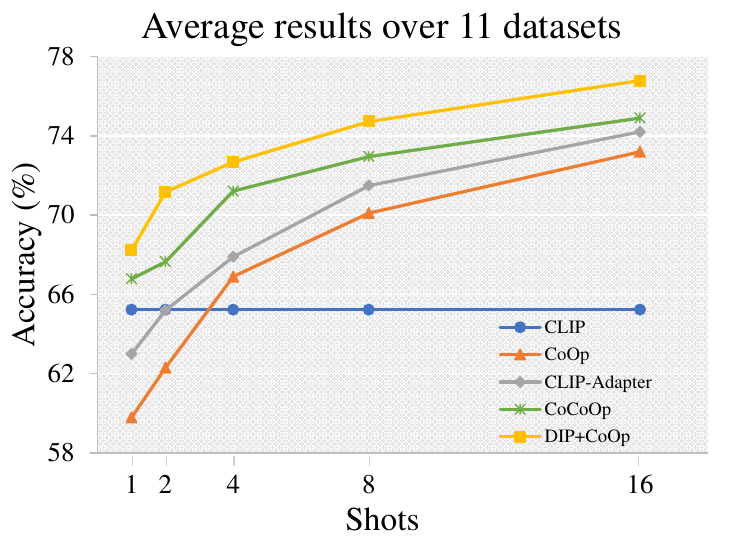}
  \setlength{\abovecaptionskip}{-0.05cm}
  \setlength{\belowcaptionskip}{-0.3cm}
  \caption{Few-shot learning Results.}
  \label{fig:fewshot}
\end{minipage}
\begin{minipage}[t]{0.55\textwidth}
  \centering
    \includegraphics[width=0.95\linewidth]{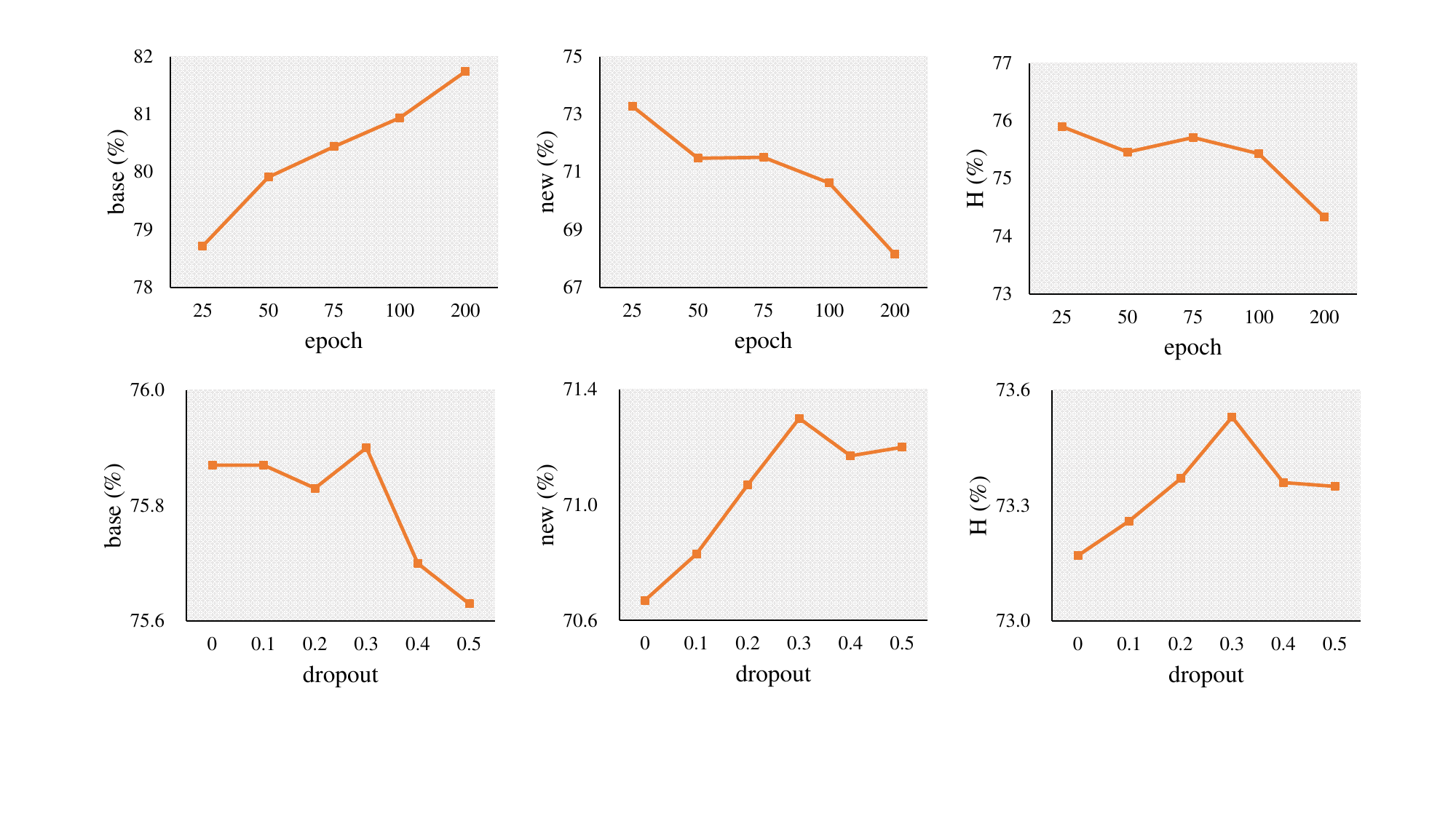}
  \setlength{\abovecaptionskip}{-0.2cm}
  \setlength{\belowcaptionskip}{-0.4cm}
  \caption{Top: Effect of training epochs. Bottom: Effect of dropout ratios.}
  \label{fig:dropout_and_epoch}
\end{minipage}
\end{figure}

In this paragraph, we will show the experiment results of DIP in the few-shot learning setting. This setting is originated from CoOp. Seen from \cref{fig:fewshot}, DIP consistently outperforms zero-shot CLIP, CoOp, and CLIP-Adapter across all the shot numbers. Such results demonstrate the superiority of DIP in adaptation ability when there are few samples in downstream tasks.

Overall, in base-to-new generalization, domain generalization, cross-dataset transfer and few-shot learning, DIP can be fruitfully integrated into existing methods and consistently reaches state-of-the-art performance while enjoying extremely high parameter efficiency.

\subsection{Analysis}
\label{subsec:analysis}

\begin{table}[tb]
    \centering   
    \scriptsize
    \setlength{\abovecaptionskip}{0.01cm}
    \setlength{\belowcaptionskip}{-0.05cm}
    \caption{Ablation study on base-to-new generalization setting}
    \setlength{\tabcolsep}{1.3mm}{
    \renewcommand{\arraystretch}{1.2}
    \begin{tabular}{|c|ccc|c|ccc|}
        \hline
        & Decomposition & Full-rank init & Dropout & \#params & base & new & H \\
        \hline
        CoOp & -  & - &  - & 2.1K & \textbf{82.69} & 63.22&71.66 \\
        \hline
        \multirow{3}{*}{DIP+CoOp}&\checkmark & - & - & 0.5K & 79.91 & 71.48 &75.46\\
        &\checkmark & \checkmark & - & 0.5K & 79.42 &73.21 &76.19\\
        &\checkmark & \checkmark &\checkmark & 0.5K & 79.70 &\textbf{73.59}& \textbf{76.53} \\
        \hline
    \end{tabular}
    }
    \label{tab:module ablation}
\end{table}

\subsubsection{Ablation study} 
We first show the impact of components of DIP step by step.

We start from transforming the text encoder. We first replace the tunable prompt tokens with our decomposed small prompt matrices, denoted as ``Decomposition'', for the CoOp method. After such replacement, the average Harmonic accuracy over 11 datasets directly improved from 71.66\% to 75.46\%. The accuracy on the base classes decreases by 2.78\% and on the new classes increases by 8.26\%. Although the adaptation ability slightly drops, the generalization ability raises quite a lot. This phenomenon proves that our low-rank design is fairly beneficial for the model's generalization ability once again. Then we integrate the special full-rank initialization into CoOp. The harmonic mean accuracy improves by 1.27\%. Finally, we add a lightweight regularization layer, Dropout. It helps us alleviate overfitting and catastrophic forgetting, resulting in further improvement.

\subsubsection{Additional experimental results}

\textbf{Effect of different training epochs} In this paragraph, we investigate that how the total training epoch could influence the adaptation result. Shown in \cref{fig:dropout_and_epoch}, we run experiments for the given epochs separately. As the training epoch increases, the accuracy on base classes continues decreasing while the accuracy on new classes continues increasing. It is reasonable because as the training continues, the model has a higher risk of forgetting its original knowledge and overfitting.

\textbf{Effect of different dropout ratios} In this paragraph, we will show the influence of different dropout ratios in DIP on ImageNet. Seen from \cref{fig:dropout_and_epoch}, as the dropout ratio increases, the base accuracy starts decreasing while the new accuracy starts increasing mostly. The harmonic mean first increases and then decreases. Dropout iss a kind of regularization, only proper regularization can help avoid overfitting and catastrophic forgetting. 

\textbf{CLIP with convolutional image encoder} In this paragraph, we show the results of DIP on CLIP with convolutional image encoder ResNet-50~\cite{he2016deep}, rather than the default ViT-B/16~\cite{dosovitskiy2020image}. Seen from \cref{tab:conv_results}, compared with baseline CoOp, DIP still largely improves the new accuracy and the harmonic mean accuracy over 11 datasets, while the base accuracy slightly drops. Compared with the latest method ProGrad, DIP shows clear superiority on base accuracy and the harmonic mean accuracy.

\begin{figure}[tb]
\centering
\setlength{\belowcaptionskip}{-0.3cm}
        \begin{minipage}{0.45\textwidth}
            \centering
            \makeatletter\def\@captype{table}\makeatother\caption{Results on ResNet-50 encoded CLIP.}
    \scriptsize
    \selectfont
    \renewcommand{\arraystretch}{1.2}
    \begin{tabular}{|c|c|c|c|c|}
        \hline
         & \#params & base & new & H \\
        \hline
        CoOp &  2.1K & \textbf{77.16} & 61.01 & 68.14 \\
        \hline
        ProGrad & 8.2K & 73.29 & \textbf{65.96} & 69.06 \\
        \hline
        DIP+CoOp & 0.5K & 75.22 & 64.98 & \textbf{69.73}\\
        \hline
    \end{tabular}
    \label{tab:conv_results}
        \end{minipage}
        \centering
        \begin{minipage}{0.45\textwidth}
        \centering
        \makeatletter\def\@captype{table}\makeatother\caption{Effect of DIP's rank on ImageNet.}
    \scriptsize
    \selectfont
    \setlength{\tabcolsep}{3mm}{
    \renewcommand{\arraystretch}{1.2}
    \begin{tabular}{|c|c|c|c|c|}
        \hline
        rank & \#params & base & new & H \\
        \hline
        1 & 0.5K & 75.87 & \textbf{70.67} & 73.17\\
        \hline
        2 & 1.0K & 76.03 & 70.13 & 72.96 \\
        \hline
        3 & 1.5K & \textbf{76.20} & 70.40 & \textbf{73.19}\\
        \hline
    \end{tabular}
    }
    \label{tab:diff_rank}
        \end{minipage}
\end{figure}

\begin{table}[tb]
    \centering
    \scriptsize
    \setlength{\belowcaptionskip}{-0.05cm}
    \caption{Results of adding DIP to image prompts on ImageNet. }
    \setlength{\tabcolsep}{2.7mm}{
    \renewcommand{\arraystretch}{1.2}
    \begin{tabular}{|c|c|c|c|c|}
        \hline
         & \#params & base & new & H \\
        \hline
        Text DIP & 0.5K & \textbf{75.87} & \textbf{70.67} & \textbf{73.17} \\
        \hline
        Image DIP & 0.5K & 74.57 & 69.40 & 71.89\\
        \hline
    \end{tabular}
    }
    \label{tab:diff_modal}
\end{table}

\textbf{Effect of different ranks} In this paragraph, we will show the influence of different ranks in DIP. Seen from \cref{tab:diff_rank}, roughly, a larger rank brings more parameters, higher base accuracy, and lower new accuracy. As a result, increasing rank is not necessarily able to improve the average accuracy.

\textbf{Results for using DIP on the image prompts} There are several works~\cite{jia2022vpt,khattak2023maple} indicating the effectiveness of the image prompts. Therefore, in this subsection, we will explore the results of applying DIP to image prompts. Seen from \cref{tab:diff_modal}, using DIP on the image side also reaches high accuracy. However, the base, new, and average accuracy of image DIP is not as good as those of text DIP. Such experiment result tells us to use DIP on the text side instead of the image side when we aim to reach high accuracy with extremely few parameters.



\section{Conclusion}

With the development of huge vision-language models, how to effectively and efficiently adapt such huge models to downstream tasks becomes a challenging problem. Much effort has been made to leverage the potential of prompt tuning in adapting vision-language models. However, existing methods suffer from inefficiency. To reach extremely efficient generalization, we propose Dense Information Prompt (DIP) based on the observation about the strong correlation between prompt rank and generalizability. Moreover, we propose a novel initialization method and a lightweight regularization module to further improve the low-rank design without adding any extra inference cost. We are the first to explore how to effectively adapt vision-language models using an extremely small number of parameters, \ie ~0.5K. Besides efficiency and effectiveness, DIP has many valuable advantages such as simplicity and robustness. Extensive experiments and analyses show the superiority of DIP sufficiently.


\bibliographystyle{unsrt}
\bibliography{neurips_2024}


\appendix
\section{Datasets}
\label{appendix_sec:dataset}

Following previous work~\cite{zhou2021coop,zhou2022conditional}, we leverage 11 image recognition datasets to verify the effectiveness of the proposed method for both the base-to-new generalization task. These datasets include two datasets for the generic object classification, \ie, ImageNet~\cite{deng2009imagenet} and Caltech101~\cite{fei2004learning}, five datasets for the fine-grained classification, \ie, OxfordPets~\cite{parkhi2012cats}, StanfordCars~\cite{krause20133d}, Flowers102~\cite{nilsback2008automated}, Food101~\cite{bossard2014food} and FGVCAircraft~\cite{maji2013fine}, one dataset for the scene recognition, \ie, SUN397~\cite{xiao2010sun}, one dataset for the action recognition, \ie, UCF101~\cite{soomro2012ucf101}, one dataset for the texture classification, \ie, DTD~\cite{cimpoi2014describing}, and one dataset for the satellite imagery recognition, \ie, EuroSAT~\cite{helber2019eurosat}. Following previous works~\cite{zhou2022conditional,zhou2021coop}, for each dataset, we split its classes equally into two non-overlapping groups, \ie, one as base classes and the other as new classes. We train all models on the base classes and perform a base/new evaluation on the base/new classes. 

For the domain generalization task, we utilize ImageNet-A~\cite{hendrycks2021IN_A}, ImageNet-R~\cite{hendrycks2021IN_R}, ImageNetv2~\cite{recht2019IN_V2} and ImageNet-S~\cite{wang2019IN_S} to verify the robustness of the model. In this setting, we need to first train the model using ImageNet, and then directly use images from other four datasets to do inference.

For the cross-dataset transfer task, the datasets are the same as those of the base-to-new generalization task. Similar to domain generalization, the model will be first trained on ImageNet and then do inference on the other 10 different datasets.

For the few-shot learning task, the datasets are the same as those of the base-to-new generalization task. The model will be trained and evaluated with 1, 2, 4, 8 and 16 shots separately. 

The dataset splitting is exactly the same as previous works~\cite{zhou2021coop,zhou2022conditional}. We report the averaged model performance over three runs with different random seeds for fair comparisons.

\section{Training Details} 
\label{appendix_sec:training_detail}
Following previous work~\cite{zhou2022conditional}, we employ ViT-B/16 as the image encoder in the CLIP. Each training image is resized to $224\times224$ before being fed into the image encoder. Some common data augmentation strategies, \eg, random crop and random flip, are used to enhance the model performance, following \cite{zhou2022conditional}. During training, we set the batch size as 32. We employ the stochastic gradient descent algorithm (SGD) to optimize the learnable parameters. As \cite{zhou2021coop}, we utilize a warm-up scheme at the first epoch, which is important for the tuning of prompts. For all the other baselines, we strictly follow the configurations of their original papers. 

To verify the effectiveness of our proposed method, we explore the improvement of integrating DIP into a lightweight prompt tuning method CoOp and a heavy prompt tuning method MaPLe separately.

For DIP+CoOp and DIP+MaPLe, we conduct a grid search to find the optimal hyper-parameters based on the configuration of CoOp and MaPLe. In the main text, we set the rank $r=1$ in DIP for all the experiments unless specially mentioned. The total number of tunable parameters in DIP+CoOp is $4*1+1*512=516$, which is merely around 0.25x of that in CoOp. For DIP+MaPLe, we also decompose its weights of the projection layer that projects text prompts to generate image prompts with rank $r_{proj}=64$. Therefore, the total number of tunable parameters in DIP+MaPLe is $(4*1+1*512)+(512*64+64*768)=82436$ per layer, and $82436*9=741924$ in total for 9 layers are modified in MaPLe by default.

\begin{table}[htbp]
    \tiny
    \setlength{\belowcaptionskip}{0.2cm}
    \setlength{\tabcolsep}{1.2mm}
    \caption{Detailed comparisons with latest methods in base-to-new generalization. H: harmonic mean~\cite{xian2017zero}. DIP+CoOp outperforms all the other lightweight methods on 8 out of 11 datasets, and DIP+MaPLe achieves better harmonic mean accuracy on 9 out of 11 datasets compared to MaPLe alone. Better still, methods modified by DIP even enjoys significantly better parameter efficiency.}
    \label{tab:appendix_results_base_to_new_generalization}
    \begin{subtable}[t]{\textwidth}
    \centering
    \caption{\textbf{Parameters comparison.}.}
    \begin{tabular}{l|cccccccc}
    \toprule
    Method & CoOp & CoCoOp & Adapter & LoRA & ProGrad & DIP+CoOp & MaPLe & DIP+MaPLe\\
    \midrule
    \#params & 2.1K & 35.4K & 525.5K & 129.0K & 8.2K & 0.5K & 3548K & 741.9K\\
    \bottomrule
    \end{tabular}
    \end{subtable}
    \begin{subtable}[t]{.3\textwidth}
    \centering
    \caption{\textbf{Average}.}
    \begin{tabular}{lcc|c}
    \toprule
    & Base & New & H \\
    \midrule
    CLIP & 69.34 & 74.22 & 71.70 \\
    CoOp  & 82.69 & 63.22 & 71.66 \\
    CoCoOp & 80.47 & 71.69 & 75.83 \\
    Adapter & 82.62 & 70.97 & 76.35 \\
    LoRA & 84.30 & 67.33 & 74.86 \\
    ProGrad & 82.79 & 68.55 & 75.00 \\
    \rowcolor{tabhighlight}
    DIP+CoOp & 80.32 & 74.73 & 77.42 \\
    MaPLe & 82.28 & 75.14 & 78.55 \\
    \rowcolor{tabhighlight}
    DIP+MaPLe & 83.17 & 75.43 & 79.11 \\
    \bottomrule
    \end{tabular}
    \end{subtable}
    \hspace{1em}
    \vspace{1em}
    \begin{subtable}[t]{.3\textwidth}
    \centering
    \caption{ImageNet.}
    \begin{tabular}{lcc|c}
    \toprule
    & Base & New & H \\
    \midrule
    CLIP & 72.43 & 68.14 & 70.22 \\
    CoOp & 76.47 & 67.88 & 71.92\\
    CoCoOp & 75.98 & 70.43 & 73.10 \\
    Adapter & 76.53 & 66.67 & 71.26 \\
    LoRA & 74.77 & 58.47 & 65.62\\
    ProGrad & 77.03 & 68.80 & 72.68 \\
    \rowcolor{tabhighlight}
    DIP+CoOp & 76.10 & 71.17 & 73.55 \\
    MaPLe & 76.66 & 70.54 & 73.47 \\
    \rowcolor{tabhighlight}
    DIP+MaPLe & 76.80 & 70.97 & 73.77 \\
    \bottomrule
    \end{tabular}
    \end{subtable}
    ~
    \begin{subtable}[t]{.3\textwidth}
    \centering
    \caption{Caltech101.}
    \begin{tabular}{l cc|c}
    \toprule
    & Base & New & H \\
    \midrule
    CLIP & 96.84 & 94.00 & 95.40 \\
    CoOp & 98.00 & 89.81 & 93.73 \\
    CoCoOp & 97.96 & 93.81 & 95.84 \\
    Adapter & 98.20 & 93.20 & 95.63 \\
    LoRA & 98.49 & 90.33 & 94.24\\
    ProGrad & 98.50 & 91.90 & 95.09 \\
    \rowcolor{tabhighlight}
    DIP+CoOp & 98.10 &95.23 &96.65 \\ 
    MaPLe & 97.74 & 94.36 & 96.02 \\
    \rowcolor{tabhighlight}
    DIP+MaPLe & 98.20 & 94.83 & 96.49 \\
    \bottomrule
    \end{tabular}
    \end{subtable}
    ~
    \begin{subtable}[t]{.3\textwidth}
    \centering
    \caption{OxfordPets.}
    \begin{tabular}{l cc|c}
    \toprule
    & Base & New & H \\
    \midrule
    CLIP & 91.17 & 97.26 & 94.12 \\
    CoOp & 93.67 & 95.29 & 94.47 \\
    CoCoOp & 95.20 & 97.69 & 96.43 \\
    Adapter & 94.40 & 94.10 & 94.25 \\
    LoRA & 94.90 & 92.57 & 93.72\\
    ProGrad & 94.40 & 95.10 & 94.75 \\
    \rowcolor{tabhighlight}
    DIP+CoOp & 95.53 & 97.93 & 96.72\\
    MaPLe & 95.43 & 97.76 & 96.58 \\
    \rowcolor{tabhighlight}
    DIP+MaPLe & 95.67 & 98.00 & 96.82 \\
    \bottomrule
    \end{tabular}
    \end{subtable}
    \hspace{1em}
    \vspace{1em}
    \begin{subtable}[t]{.3\textwidth}
    \centering
    \caption{StanfordCars.}
    \begin{tabular}{l cc|c}
    \toprule
    & Base & New & H \\
    \midrule
    CLIP & 63.37 & 74.89 & 68.65 \\
    CoOp & 78.12 & 60.40 & 68.13 \\
    CoCoOp & 70.49 & 73.59 & 72.01 \\
    Adapter & 77.13 & 69.23 & 72.97 \\
    LoRA & 81.07 & 65.30 & 72.34 \\
    ProGrad & 79.00 & 67.93 & 73.05 \\
    \rowcolor{tabhighlight}
    DIP+CoOp & 71.33  & 74.47 & 72.87 \\
    MaPLe & 72.94 & 74.00 & 73.47 \\
    \rowcolor{tabhighlight}
    DIP+MaPLe & 76.00 & 73.00 & 74.47 \\
    \bottomrule
    \end{tabular}
    \end{subtable}
    ~
    \begin{subtable}[t]{.3\textwidth}
    \centering
    \caption{Flowers102.}
    \begin{tabular}{l cc|c}
    \toprule
    & Base & New & H \\
    \midrule
    CLIP & 72.08 & 77.80 & 74.83 \\
    CoOp & 97.60 & 59.67 & 74.06 \\
    CoCoOp & 94.87 & 71.75 & 81.71 \\
    Adapter & 97.70 & 70.83 & 82.13 \\
    LoRA & 98.23 & 60.20 & 74.65\\
    ProGrad & 96.27 & 71.07 & 81.77 \\
    \rowcolor{tabhighlight}
    DIP+CoOp & 95.53 & 75.23 & 84.18\\
    MaPLe & 95.92 & 72.46 & 82.56 \\
    \rowcolor{tabhighlight}
    DIP+MaPLe & 97.00 & 73.77 & 83.80 \\
    \bottomrule
    \end{tabular}
    \end{subtable}
    ~
    \begin{subtable}[t]{.3\textwidth}
    \centering
    \caption{Food101.}
    \begin{tabular}{l cc|c}
    \toprule
    & Base & New & H \\
    \midrule
    CLIP & 90.10 & 91.22 & 90.66 \\
    CoOp & 88.33 & 82.26 & 85.19 \\
    CoCoOp & 90.70 & 91.29 & 90.99 \\
    Adapter & 90.40 & 90.40 & 90.40 \\
    LoRA & 88.57 & 87.30 & 87.93\\
    ProGrad & 90.17 & 89.53 & 89.85 \\
    \rowcolor{tabhighlight}
    DIP+CoOp & 90.83  & 92.03 & 91.43\\
    MaPLe & 90.71 & 92.05 & 91.38 \\
    \rowcolor{tabhighlight}
    DIP+MaPLe & 90.63 & 91.90 & 91.26 \\
    \bottomrule
    \end{tabular}
    \end{subtable}
    \hspace{1em}
    \vspace{1em}
    \begin{subtable}[t]{.3\textwidth}
    \centering
    \caption{FGVCAircraft.}
    \begin{tabular}{l cc|c}
    \toprule
    & Base & New & H \\
    \midrule
    CLIP & 27.19 & 36.29 & 31.09 \\
    CoOp & 40.44 & 22.30 & 28.75 \\
    CoCoOp & 33.41 & 23.71 & 27.74 \\
    Adapter & 39.57 & 32.27 & 35.55 \\
    LoRA & 46.27 & 28.83 & 35.53\\
    ProGrad & 42.63 & 26.97 & 33.04 \\
    \rowcolor{tabhighlight}
    DIP+CoOp & 35.17 & 35.57 & 35.37  \\
    MaPLe & 37.44 & 35.61 & 36.50 \\
    \rowcolor{tabhighlight}
    DIP+MaPLe & 39.03 & 37.17 & 38.08 \\
    \bottomrule
    \end{tabular}
    \end{subtable}
    ~
    \begin{subtable}[t]{.3\textwidth}
    \centering
    \caption{SUN397.}
    \begin{tabular}{l cc|c}
    \toprule
    & Base & New & H \\
    \midrule
    CLIP & 69.36 & 75.35 & 72.23 \\
    CoOp & 80.60 & 65.89 & 72.51 \\
    CoCoOp & 79.74 & 76.86 & 78.27 \\
    Adapter & 81.67 & 73.93 & 77.61 \\
    LoRA & 79.73 & 69.00 & 73.98\\
    ProGrad & 80.70 & 71.03 & 75.56 \\
    \rowcolor{tabhighlight}
    DIP+CoOp & 79.60 & 77.80 & 78.69 \\
    MaPLe & 80.82 & 78.70 & 79.75 \\
    \rowcolor{tabhighlight}
    DIP+MaPLe & 81.27 & 78.13 & 79.67 \\
    \bottomrule
    \end{tabular}
    \end{subtable}
    ~
    \begin{subtable}[t]{.3\textwidth}
    \centering
    \caption{DTD.}
    \begin{tabular}{lcc|c}
    \toprule
    & Base & New & H \\
    \midrule
    CLIP & 53.24 & 59.90 & 56.37 \\
    CoOp & 79.44 & 41.18 & 54.24 \\
    CoCoOp & 77.01 & 56.00 & 64.85 \\
    Adapter & 80.47 & 52.23 & 63.35 \\
    LoRA & 82.93 & 54.90 & 66.06\\
    ProGrad & 76.70 & 46.67 & 58.03 \\
    \rowcolor{tabhighlight}
    DIP+CoOp & 75.67 & 60.43 & 67.20 \\
    MaPLe & 80.36 & 59.18 & 68.16 \\
    \rowcolor{tabhighlight}
    DIP+MaPLe & 82.10 & 59.50 & 69.00 \\
    \bottomrule
    \end{tabular}
    \end{subtable}
    \hspace{2.1em}
    \begin{subtable}[t]{.3\textwidth}
    \centering
    \caption{EuroSAT.}
    \begin{tabular}{lcc|c}
    \toprule
    & Base & New & H \\
    \midrule
    CLIP & 56.48 & 64.05 & 60.03 \\
    CoOp & 92.19 & 54.74 & 68.69 \\
    CoCoOp & 87.49 & 60.04 & 71.21 \\
    Adapter & 86.93 & 64.20 & 73.86 \\
    LoRA & 94.90 & 65.67 & 77.62\\
    ProGrad & 91.37 & 56.53 & 69.85 \\
    \rowcolor{tabhighlight}
    DIP+CoOp & 82.70 & 65.40 & 73.04 \\
    MaPLe & 94.07 & 73.23 & 82.35 \\
    \rowcolor{tabhighlight}
    DIP+MaPLe & 93.67 & 74.33 & 82.89 \\
    \bottomrule
    \end{tabular}
    \end{subtable}
    \hspace{2em}
    \begin{subtable}[t]{.3\textwidth}
    \centering
    \caption{UCF101.}
    \begin{tabular}{lcc|c}
    \toprule
    & Base & New & H \\
    \midrule
    CLIP & 70.53 & 77.50 & 73.85 \\
    CoOp & 84.69 & 56.05 & 67.46 \\
    CoCoOp & 82.33 & 73.45 & 77.64 \\
    Adapter & 85.80 & 73.63 & 79.25 \\
    LoRA & 87.47 & 68.03 & 76.53\\
    ProGrad & 83.90 & 68.50 & 75.42 \\
    \rowcolor{tabhighlight}
    DIP+CoOp & 82.93 & 76.70 & 79.69 \\
    MaPLe & 83.00 & 78.66 & 80.77 \\
    \rowcolor{tabhighlight}
    DIP+MaPLe & 84.50 & 78.13 & 81.19 \\
    \bottomrule
    \end{tabular}
    \end{subtable}
\end{table}

\section{Competitors}
\begin{enumerate}
    \item \textbf{CLIP}~\cite{radford2021learning}: CLIP is a strong baseline vision-language model that is pre-trained on a large number of image-text pairs from the web by learning a contrastive objective. CLIP enables strong zero-shot adaptation ability on various downstream tasks by using fixed text prompts, \ie \textit{a photo of a}.
    \item \textbf{CoOp}~\cite{zhou2021coop}: CoOp replaces the fixed text prompts in CLIP with tunable text prompts to improve the adaptation ability of the vision-language model. CoOp shows excellent performance in few-shot situations.
    \item \textbf{CoCoOp}~\cite{zhou2022conditional}: CoCoOp replaces the isolated tunable text prompts in CoOp with conditional text prompts, which receive extra gradients from the image features besides text features. CoCoOp largely improves the generalization ability of the vision-language model, getting good results on base-to-new generalization and domain adaptation.
    \item \textbf{CLIP-Adapter}~\cite{gao2021clip-adapter}: CLIP-Adapter adopts the thoughts of classic Adapter~\cite{houlsby2019parameter} to use serial linear layers and activation functions to adapt for downstream tasks. It is simple yet effective in few-shot learning.
    \item \textbf{LoRA}~\cite{hu2021lora}: LoRA adopts low-rank decomposition for weights in FC layers. It is efficient and earns good results in NLP field.
    \item \textbf{MaPLe}~\cite{khattak2023maple}: MaPLe simultaneously adds prompts to the image encoder and text encoder of CLIP. To trigger more information exchange between the image side and text side, MaPLe generate image prompts from the projection of text prompts. Though effective, such design brings quite heavy cost.  
    \item \textbf{ProGrad}~\cite{zhu2023prograd}: ProGrad only updates the text and image prompts whose gradient are aligned (or non-conflicting) to the general knowledge, which is represented as the optimization direction offered by the pre-defined prompt predictions. Such regularization helps it finish good adaptation and generalization.
\end{enumerate}

\begin{table}[tb]
    \centering
    \scriptsize
    \setlength{\belowcaptionskip}{-0.05cm}
    \caption{Base-to-new generaliation performances based on SLIP~\cite{mu2022slip}. }
    \setlength{\tabcolsep}{2.7mm}{
    \renewcommand{\arraystretch}{1.2}
    \begin{tabular}{|c|c|c|c|c|}
        \hline
         & \#params & base & new & H \\
        \hline
        CoOp & 2.1K & \textbf{68.45} & 42.77 & 52.64 \\
        \hline
        DIP+CoOp & 0.5K & 62.53 & \textbf{47.78} & \textbf{54.17}\\
        \hline
    \end{tabular}
    }
    \label{tab:slip}
\end{table}

\begin{table}[tb]
    \centering
    \fontsize{9}{11.25}
    \selectfont
    \caption{Results of different combinations of learning rates and weight decays under 16-shot learning setting on ImageNet}
    \setlength{\tabcolsep}{1.3mm}{
    \renewcommand{\arraystretch}{1.2}
    \begin{tabular}{|c|c|c|c|}
        \hline
        \diagbox{lr}{Acc}{wd}& 1e-4 & 5e-4 & 1e-3 \\
        \hline
        1e-3 &70.73  & 70.67 & 70.78\\
        \hline
        2e-3 & 70.80& 70.83 &70.77 \\
        \hline
        3e-3& 70.83& 70.83 &  70.81\\
        \hline
    \end{tabular}
    }
    \label{tab:lr_wd_comb}
\end{table}

\begin{table}[tb]
    \centering
    \scriptsize
    \setlength{\belowcaptionskip}{-0.05cm}
    \caption{Deep prompts in different depths for DIP+CoOp.}
    \setlength{\tabcolsep}{2.7mm}{
    \renewcommand{\arraystretch}{1.2}
    \begin{tabular}{|c|c|c|c|c|}
        \hline
        depth & \#params & base & new & H \\
        \hline
        1 & 0.5K & 76.37 & 74.69 & 75.52 \\
        \hline
        2 & 1.0K & 77.81 & 73.87 & 75.79\\
        \hline
        3 & 1.5K & 79.55 & 72.72 & 75.98 \\
        \hline
        4 & 2.0K & 80.07 & 72.69 & 76.20\\
        \hline
        5 & 2.6K & 80.24 & 73.37 & 76.65 \\
        \hline
        6 & 3.1K & 80.64 & 73.39 & 76.85\\
        \hline
    \end{tabular}
    }
    \label{tab:deepprompt}
\end{table}

\section{Complete Results in Base-To-New Generalization}
As shown in \cref{tab:appendix_results_base_to_new_generalization}, overall, DIP shows quite competitive performance, and the superiority mainly relies on the improvement of new class accuracy. In other words, DIP largely improves the generalization ability of CLIP. In particular, compared with CoOp, DIP+CoOp gets $10.95$\% accuracy gain on the new classes and $2.83$\% accuracy drop on the base classes, showing strongest harmonic mean accuracy on 8 out of 11 datasets among all the lightweight existing methods. Compared with a heavy method MaPLe, DIP+MaPLe achieves better harmonic mean accuracy on 9 out of 11 datasets.

\section{Experiments on a Different Vision-Language Architecture}
In this paragraph, we show the results of DIP on another vision-language architecture, SLIP~\cite{mu2022slip}, besides CLIP. Seen from \cref{tab:slip}, DIP+CoOp earns much higher new accuracy and harmonic mean accuracy than the original CoOp. 

\section{Effect of Deep Prompts}
In this paragraph, we extend the shallow prompts in DIP+CoOp to the deep prompts. We record the accuracy change as we increase the layers including tunable prompts, following the last equation in Section 3.1 in the main text. Results are shown in \cref{tab:deepprompt}. As the depth increases, the base accuracy keeps increasing while the new accuracy first decreases and then increases. Overall, more depth generally leads to higher harmonic mean accuracy. Therefore, it is possible to further improve the performance of our method by increasing the prompt depth.

\section{Effect of Different learning rates and Weight Decays}
In this paragraph, we investigate the effect of normal hyper-parameters learning rate and weight decay. We wonder if tuning them carefully would lead to significant improvement. Seen from \cref{tab:lr_wd_comb}, the performance of DIP stays stable whatever the learning rate and weight decay vary. The conclusion here is that our method DIP is robust for learning rate and weight decay.

\begin{table}[tb]
    \centering
    \scriptsize
    \setlength{\belowcaptionskip}{-0.05cm}
    \caption{Training, storage, and inference efficiencies. }
    \renewcommand{\arraystretch}{1.2}
    \begin{tabular}{|c|c|c|c|}
        \hline
         & \#params & \makecell[c]{Training\\throughput} & \makecell[c]{Inference\\throughput} \\
        \hline
        CoOp & 2.1K & \textbf{93} image/s & \textbf{738} images/s \\
        \hline
        CoCoOp & 35.4K & 5 images/s & 13 images/s \\
        \hline
        ProGrad & 8.2K & 56 images/s & 732 images/s \\
        \hline
        DIP+CoOp & \textbf{0.5K} & 91 images/s & \textbf{738} images/s\\
        \hline
    \end{tabular}
    \label{tab:speed}
\end{table}

\section{Efficiency Comparison} 
In this paragraph, we will give a comprehensive analysis of all the training, storage, and inference efficiencies for DIP and several existing methods. Since the parameter scale of CLIP-Adapter~\cite{gao2021clip-adapter} is significantly larger than others, 
we do not contain CLIP-Adapter into comparison. 

For a fair comparison, we do all the speed tests on the same GPU. Results are shown in \cref{tab:speed}. Our proposed DIP shares nearly the same fastest training and inference speeds with the simplest method CoOp, and more importantly, DIP merely uses 0.5K parameters, which is a lot more storage-efficient than other methods. Specially, compared with classic method CoCoOp, DIP enjoys \textbf{$>$18x} training speed, \textbf{$>$56x} inference speed and \textbf{$<$70x} storage usage. Besides complex structures, the huge gap in the inference speed is partly owing to the huge memory cost of CoCoOp, which forces us to adopt a smaller batch size than other methods for CoCoOp. Moreover, compared with the latest method ProGrad, DIP enjoys \textbf{$>$1.6x} training speed, comparable inference speed, and \textbf{$<$60x} storage usage, which adequately demonstrates the super efficiency of DIP. 


\end{document}